\documentclass[a4paper, 10pt, conference]{IEEEtran}
\IEEEoverridecommandlockouts

\usepackage{cite}
\usepackage{amsmath,amssymb,amsfonts}
\usepackage{algorithmic}
\usepackage{graphicx}
\usepackage{textcomp}

\usepackage{romannum}
\usepackage[pagebackref=true,breaklinks=true,letterpaper=true,colorlinks,bookmarks=false]{hyperref}

\usepackage{xcolor}
\def\BibTeX{{\rm B\kern-.05em{\sc i\kern-.025em b}\kern-.08em
    T\kern-.1667em\lower.7ex\hbox{E}\kern-.125emX}}
\begin{document}

\title{OTPose: Occlusion-Aware Transformer for Pose\\Estimation in Sparsely-Labeled Videos}

\author{\IEEEauthorblockN{Kyung-Min Jin}
\thanks{
This work was supported by Institute of Information \& communications Technology Planning \& Evaluation (IITP) grant funded by the Korea government (MSIT) (No. B0101-15-0266, Development of High Performance Visual BigData Discovery Platform for Large-Scale Realtime Data Analysis, No. 2021-0-02068, Artificial Intelligence Innovation Hub).}

\IEEEauthorblockA{\textit{Dept. Artificial Intelligence} \\ 
\textit{Korea University}\\
Seoul, South Korea \\
km\_jin@korea.ac.kr}
\and
\IEEEauthorblockN{Gun-Hee Lee}
\IEEEauthorblockA{\textit{Dept. Computer Engineering} \\
\textit{Korea University}\\
Seoul, South Korea \\
gunhlee@korea.ac.kr}
\and
\IEEEauthorblockN{Seong-Whan Lee}
\IEEEauthorblockA{\textit{Dept. Artificial Intelligence} \\
\textit{Korea University}\\
Seoul, South Korea \\
sw.lee@korea.ac.kr
}}

\maketitle

\begin{abstract}

Although many approaches for multi-human pose estimation in videos have shown profound results, they require densely annotated data which entails excessive man labor. Furthermore, there exists occlusion and motion blur that inevitably lead to poor estimation performance. To address these problems, we propose a method that leverages an attention mask for occluded joints and encodes temporal dependency between frames using transformers. First, our framework composes different combinations of sparsely annotated frames that denote the track of the overall joint movement. We propose an occlusion attention mask from these combinations that enable encoding occlusion-aware heatmaps as a semi-supervised task. Second, the proposed temporal encoder employs transformer architecture to effectively aggregate the temporal relationship and keypoint-wise attention from each time step and accurately refines the target frame's final pose estimation. We achieve state-of-the-art pose estimation results for PoseTrack2017 and PoseTrack2018 datasets and demonstrate the robustness of our approach to occlusion and motion blur in sparsely annotated video data.
\end{abstract}


\section{Introduction}

Human pose estimation in images has been studied extensively, and methods \cite{lee2021uncertainty,yang2007reconstruction,hrnet,cpm,transpose,ssdpose,bodyjoints}
have shown remarkable results with the advent of deep learning. However, when applied to a video, these single image-based methods tend to produce temporally inconsistent (jitter or unsmooth) results. Meanwhile, there are several problems with video pose estimation. First, large benchmark datasets are essential to achieve high performance, but the extensive cost of dense annotation makes it challenging to increase the dataset size. Annotations for every frame may also be meaningless because poses are not significantly different unless large motions exist between frames. Second, inevitable problems such as occlusion, motion blur, and video defocus still exist in videos.

Thus, current methods\cite{posewarper,dcpose,3dhrnet} for videos have been studied to capture temporal correlations between frames. With great success in image classification using convolutional neural networks (CNNs) \cite{lee2003pattern,new1,lee1990translation}, many approaches are based on CNNs.
The basic idea of CNN-based multi-human pose estimation methods for video is capturing each keypoint relationship between frames representing a temporal dependency. However, implicitly encoding keypoint relationships results in inaccurate predictions for keypoints with severe occlusion or motion blur. In addition, CNN-based methods have a fundamental problem: the receptive field is limited, so the depth of the network must be linearly increased to model long-range dependency between all pixel pairs in consecutive frames.

As another line of work, LPM \cite{lpm} extends an image-based method CPM \cite{cpm} by employing recurrent neural networks (RNNs) such as long short term memory (LSTM) to intuitively capture temporal dependency. RNN-based approaches perform well in single human pose estimation, which only contains spatially sparse scenes. However, they generally fail in multi-human pose estimation, a task with many occlusion and crowded scenes.

\begin{figure}[t]
\centering
\includegraphics[width=\linewidth]{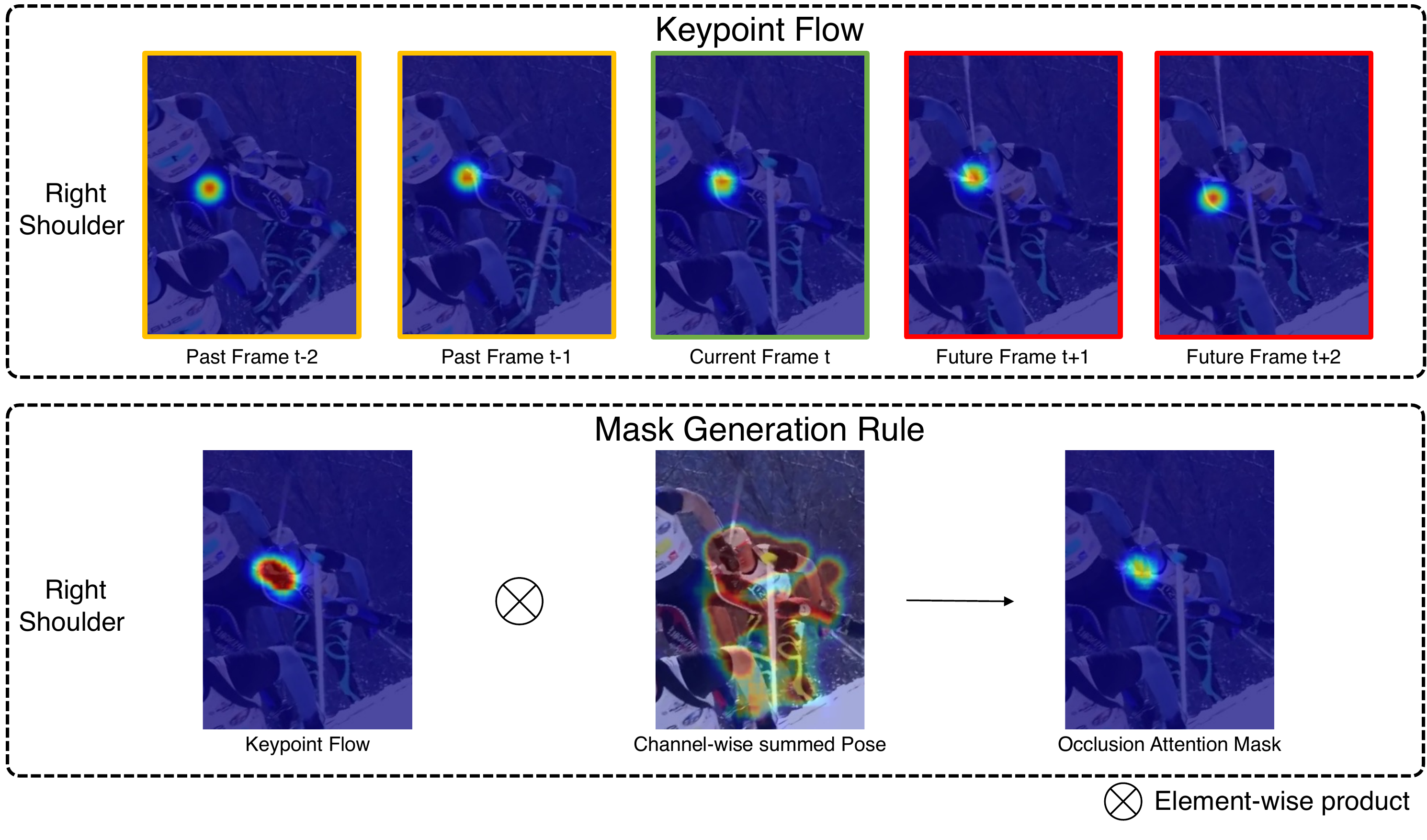}
\caption{Visualization of occlusion attention mask. This mask can effectively limit the candidate position for keypoint prediction by element-wise product with the track of total joints. This operation helps to predict joints by focusing on severely occluded areas.}
\label{fig:generation rule of occlusion attention mask.}
\end{figure}

\begin{figure*}[t]
\centering
\includegraphics[width=\textwidth]{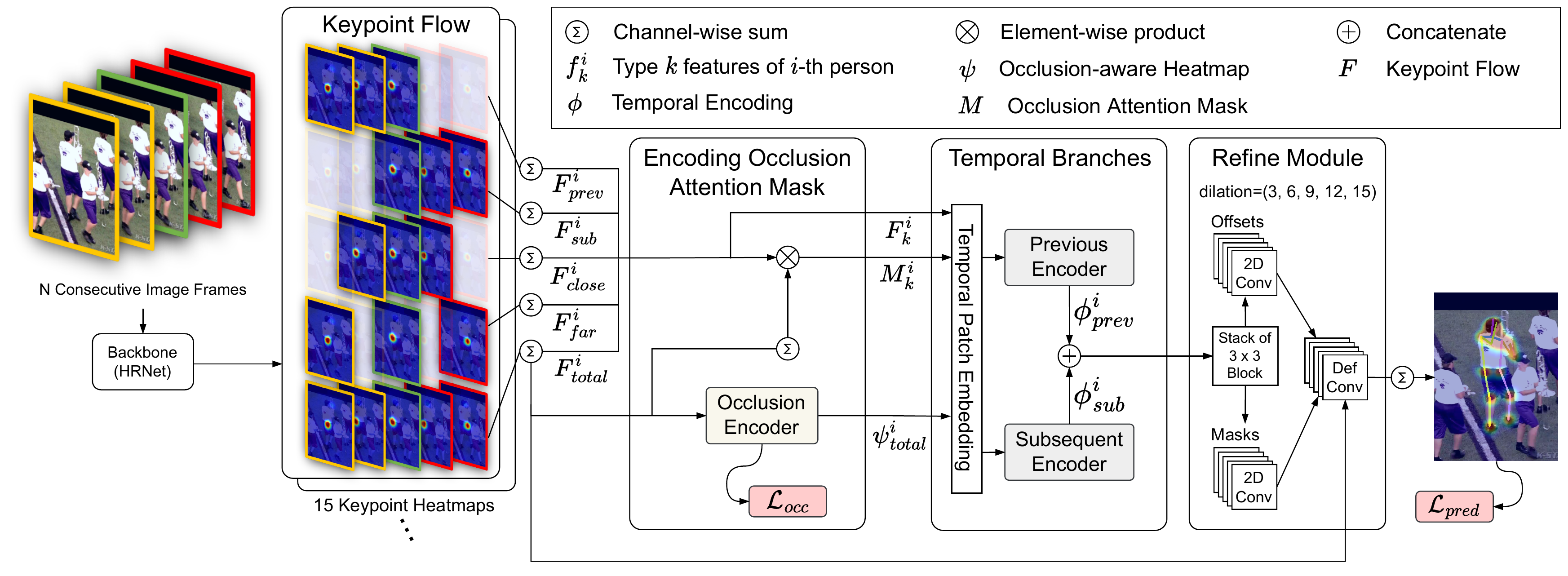}
\caption{The overall architecture of our OTPose. First, occlusion-aware heatmaps are processed from keypoint flow that denotes the track of each joint. Then, keypoint flows, masks, and occlusion-aware heatmaps are divided into two temporal branches and independently encoded using a transformer encoder. Each type $k$ denotes previous (prev), subsequent (sub), close, far, and total.}
\label{fig:pipeline}
\end{figure*}

In this paper, we propose a network named OTPose (Occlusion-aware Transformer for Pose estimation) that explicitly encodes occlusion using a mask as a semi-supervised task and intuitively understands temporal dependency using a transformer encoder. First, we effectively accumulate occlusion-specific features and introduce an attention mask that focuses on the overlapped area of easily occluded keypoints such as wrists and ankles. As visualized in Fig.~\ref{fig:generation rule of occlusion attention mask.}, occlusion attention masks enable us to encode robust features for occlusion. The mask made this possible by increasing the activation value in the keypoint heatmap area that has low confidence. Second, two temporal-domain branches independently encode unique pose features that focus on past and future frames, respectively. Furthermore, we employ a temporal encoder for each branch to overcome temporal-dependent issues such as occlusion and motion blur from various keypoint flows with efficient keypoint-wise attention. Finally, the keypoint positions are refined using the difference between these branches. In summary, the main contributions of the paper are: 

\begin{itemize}
    \item We propose a novel approach OTPose for estimating 2D poses in videos that effectively aggregates keypoint-wise attention with two temporal-domain branches, parameterized as transformer encoders.
    \item We propose an occlusion attention mask with increased activation values at keypoint locations that are easily occluded and use them to encode occlusion-aware heatmaps as a semi-supervised task. 
    \item Our method jointly optimizes pose estimation and a semi-supervised task to reduce the reliance on dense annotations. We achieve state-of-the-art on large benchmark datasets: PoseTrack2017 and PoseTrack2018.
\end{itemize}

\section{Related Work}

\subsection{Multi-Person Pose Estimation in Video}
Many pose estimation approaches in images \cite{hrnet,cpm,transpose,ssdpose,bodyjoints,hur2012supervised} are inappropriate to be directly applied to the video because they do not consider a temporal relationship. Thus, advances have been made in videos by expanding the image-based methods: By adding RNNs to CPM \cite{cpm}, LPM \cite{lpm} intuitively captures the temporal dependency between keypoint positions in each frame, and 3D HRNet \cite{3dhrnet} replaces CNNs of HRNet \cite{hrnet} with 3D CNNs. Recently, modern approaches generally estimate pose per frame with an image-based backbone network and then refine it through temporal information. PoseWarper \cite{posewarper} learns the warping mechanism through label propagation in sparsely labeled videos \cite{posetrack2018}. In addition, DCPose \cite{dcpose} proposes refining a pose using bidirectional frames. 

These prior works have shown remarkable performances by introducing temporal concepts. However, implicit learning a problem, such as occlusion and motion blur, results in a low estimation quality in complex scenes. In contrast to these methods, we explicitly train our model using occlusion attention masks.

\subsection{Transformer}

As frameworks have evolved from machine learning \cite{xi2002facial,lim2000text} to deep learning, CNNs have been mainly used in the computer vision field. However, with the success of the attention-based method in NLP, transformers combined with CNNs have brought inspiration for new approaches \cite{vivit,vit,transpose,3dspatiotemporal} to this field. Because the essential operation of the transformer is multi-head self-attention, which shows superior performance in modeling long-range dependencies, it can be effectively used to focus on spatio-temporal relations. After ViT \cite{vit}, based on a pure-transformer, successfully outperforms CNNs counterparts in classification performance on large image datasets, several methods using transformers also emerged in pose estimation in images. Transpose \cite{transpose} proposes a method that creates interpretability and captures long-range dependency. \cite{3dspatiotemporal} also encodes 3D information using a spatial transformer with a temporal transformer. Unlike these studies, we extend transformers to predict the occluded keypoints distribution and heatmaps representing the spatio-temporal distribution of keypoints for 2D video human pose estimation.

\section{Method}

Our main objective is to build a model that leverages sparse labels without being affected by occlusion or motion blur. We introduce each component of OTPose in detail below. 

\subsection{Encoding Occlusion Attention Mask}
\subsubsection{\textbf{Occlusion Attention Mask}}

We propose an occlusion attention mask from a keypoint flow $F$ that denotes a track of the movement of joints between sampled frames to address the fundamental problems of pose estimation in videos, such as occlusion and motion blur. We can obtain an explicit mask by element-wise product that increases the activation value in areas where occlusion or motion blur occurs, while the probability values in other areas are reduced. As visualized in Fig.~\ref{fig:generation rule of occlusion attention mask.}, masks focus on unpredictable joints due to severe occlusion or motion blur.

We consider five combinations of frame heatmaps; previous, subsequent, close, far, and total types, as shown in Fig.~\ref{fig:pipeline}. The past and future frames heatmaps are penalized using a distance away from the current frame when generating a keypoint flow, except for the total type. This is given by,
\begin{equation}
    \hat{{H}}_{t\pm d}^i = \frac{1}{d+1}{H}_{t\pm d}^i,
\end{equation}
where $\hat{{H}}_{t\pm d}^i$ denotes penalized heatmaps of the frame $t$ for a person $i$ by a frame distance $d\in\{1,2\}$. A previous type aggregates penalized past frame heatmaps $\hat{{H}}_{t-d}^i$ with current frame heatmaps ${H}_t^i$, and the subsequent type uses penalized future frame heatmaps. We use heatmaps of the same distance in each past and future domain in the close and far types. Then attention mask is generated from each keypoint flow type by an element-wise product with a channel-wise sum of ${F}_{total}$. This can be formulated as follows:
\begin{equation}
  {M}_k^i = {F}_k^i \otimes \sum_{j=1}^{N_j}{F}_{total}^{i,j},
\end{equation}
where $\otimes$, $N_j$, and ${M}_k^i$ denote an element-wise product between each keypoint flow type and the channel-wise sum of ${F}_{total}$, the number of joints, and an occlusion attention mask for keypoint flow type $k$, respectively.

\begin{figure}[t]
\centering
\includegraphics[width=0.9\linewidth]{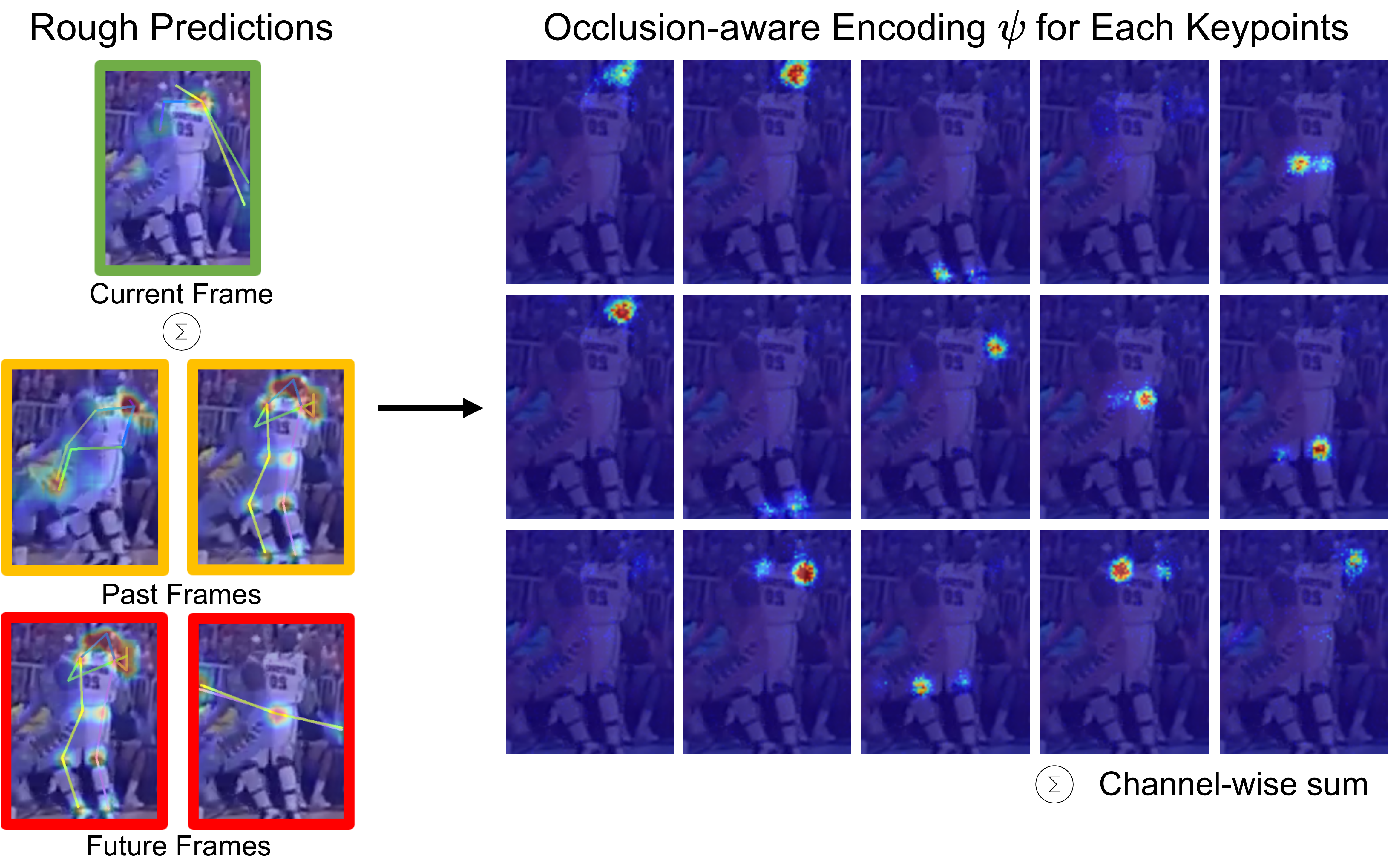}
\caption{We show the necessity for occlusion encoder. If the quality of backbone predictions deteriorates due to severe occlusion and motion blur, the attention mask may not be able to focus on the occluded area. Occlusion encoder can generalize joint position even when rough prediction quality is poor.}
\label{fig:Occlusion Encoder}
\end{figure}

\subsubsection{\textbf{Semi-supervised Occlusion Encoder}}
We design a semi-supervised task to generate occlusion-aware heatmaps $\psi$ from sampled frames. Occlusion encoder (OE) uses the temporal relationship of keypoint flow to predict occluded or blurred keypoints.
\begin{equation}
  \psi_{total}^i = \operatorname{OE}({F}_{total}^i), \quad \hat{G}: \sigma({M}_{total}^i + {G^i})
\end{equation}
where $\psi_{total}^i$ and $\hat{G}$ denote occlusion-aware heatmaps and a pseudo label: a sum of ground truth $G$ with mask $M$ normalized by $\sigma$. As visualized in Fig.~\ref{fig:Occlusion Encoder}, occlusion encoder can recover lost information through $\hat{G}$ due to occlusion, motion blur, or inaccurate backbone predictions in a particular frame.

\subsection{Temporal Branch}

\begin{figure}[t]
\begin{center}
\includegraphics[width=0.9\linewidth]{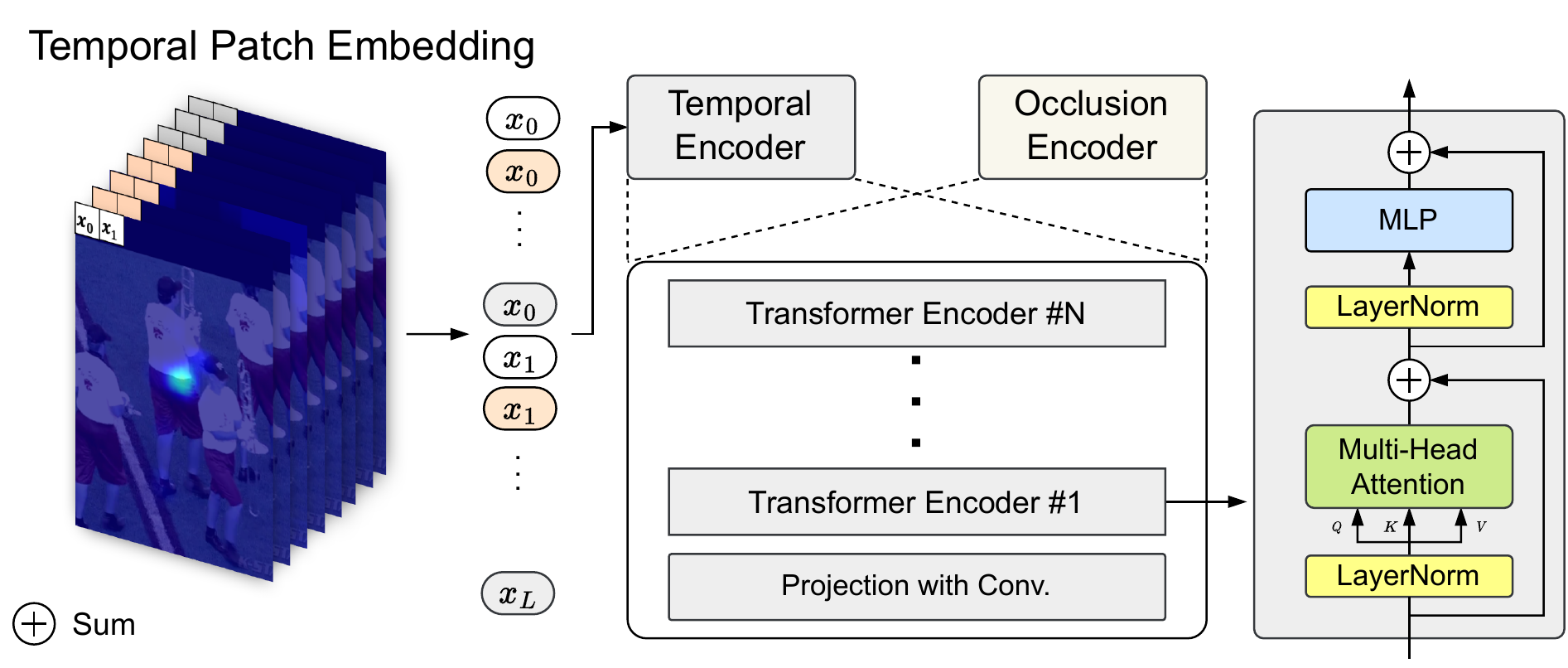}
\caption{Temporal patch embedding contains temporally different flows of each keypoint and global relation between them at the identical patch location that spans input volume. Transformer encoder layers iteratively capture temporal dependencies by self-attention.}
\label{fig:Temporal Patch Embedding}
\end{center}
\end{figure}

After masks ${M}_k^i$ and occlusion-aware heatmaps $\psi_{total}^i$ are obtained, two temporal branches encode features using the transformer encoder as showed in Fig.~\ref{fig:Temporal Patch Embedding}, representing past and future frame information. Our Temporal encoder (TE) divides and learns features since the difference between two estimated poses is refined through deformable convolution (in section~\Romannum{3}-C) that samples different temporally activated regions. At each branch, we process the corresponding stacked features. It can be expressed as:
\begin{equation}
    \begin{aligned}
        &{\phi_{prev}^i} = \operatorname{TE}(C^i;  {M}_{prev}^i; F_{prev}^i), \\
        &{\phi_{sub}^i} = \operatorname{TE}(C^i; {M}_{sub}^i; F_{sub}^i), \\
        &C^i =\left\{M_k^i; F_k^i; M_{total}^i; \psi_{total}^i \mid k \in \{close, far\}\right\}.
  \end{aligned}
\end{equation}
Features of each branch independently pass through the transformer encoder, and the head generates heatmaps $\phi_{prev}^i$ and $ \phi_{sub}^i$ for each branch. Since the max activation position of the heatmap is predicted as a keypoint location, we do not use the decoder. 

\subsection{Refine Module}
Given the output of the temporal encoder, refine module that exploits the deconvolution network \cite{dcnv2} computes offsets and masks between branches. First, with $F_{total}$, $\phi_{prev}^i$ and $\phi_{sub}^i$ are processed by a stack of $3 \times 3$ convolution layers. Then independent 2D convolutions compute the offsets and mask at each dilation $d=\{3,6,9,12,15\}$. Small dilation captures local information where the motion is small, while large dilation encodes fast motion. Then the refine module independently predicts the position from keypoint flow with offsets and masks in parallel. Finally, the output from each dilation is added and normalized to refine keypoint locations.

\subsection{Encoder}
\subsubsection{\textbf{Temporal Patch Embedding}}
Inspired by ViViT \cite{vivit}, we implement a temporal patch embedding by stacking features, as shown in Fig.~\ref{fig:Temporal Patch Embedding}. First, a heatmap $\mathbf{X}=\{\mathbf{x}_i\mid \mathbf{x}_i \in R^{H\times W}, i\in\{1, 2, \ldots, N_j\}\}$ is divided into a grid of patches $\frac{H}{P_h}\times\frac{W}{P_w}$, where $N_j$ denotes the number of joints. Then each patch is stacked at the same location to form $D$-dimensional embedded feature volume $\mathbf{Z}\in\mathbb{R}^{D \times L}$, where $D$ and $L$ denote $N_j\times N_f$ and $\frac{H}{P_h}\times\frac{W}{P_w}$. Here, $N_f$ denotes the number of stacked features. We set $N_f$ as 8 for temporal encoder and 1 for occlusion encoder. By embedding the patches in this way, temporal information accumulates into each patch at the same location. $\mathbf{Z}$ is then projected as $\mathbf{Q}, \mathbf{K}$, and $\mathbf{V}$ to represent query, key, and value. 

\subsubsection{\textbf{Attention Mechanism}} 
We can capture the pairwise relationship of each joint by self-attention. Because we employ a temporal patch embedding, estimating performance in videos can be maximized by attention between joints. Computed attention maps can be regarded as a contribution of other keypoints when predicting the target keypoint. Attention operation $\mathbf{A}\in\mathbb{R}^{D\times D}$ is defined as:
\begin{equation}
  \mathbf{A} = softmax(\frac{\mathbf{Q}\mathbf{K}^T}{\sqrt{D}}).
\end{equation}
Patch-level attention has a complexity of $O(L^2)$, which has extensive cost for a high-resolution image in both memory and computation. However, our keypoint-wise attention has an efficient memory and computation complexity $O(D^2)$ while maintaining or improving performances. Each query $\mathbf{q}_i \in \mathbb{R}^{L}$ of token $\mathbf{x}_i\in\mathbb{R}^{L}$ for keypoint $i$ computes similarities with all keys to obtain $\mathbf{A}_{i,:}\in\mathbb{R}^{1\times D}$. It determines how much the query depends on other tokens. Further, it indicates how much the query keypoint refers to other token keypoints when estimating the keypoint location. Then $\mathbf{A}$ is multiplied by $\mathbf{V}$ and added to $\mathbf{Z}$.

Here, LayerNorm (LN) is applied before every multi-head self-attention (MSA) and MLP block. GELU is used for the MLP activation function. The embedded feature volume $\mathbf{Z}$ is processed as:
\begin{equation}
    \begin{aligned}
        &\overline{\mathbf{Z}}^{\ell}=\alpha^{\ell} \operatorname{MSA}(\operatorname{LN}(\mathbf{Z}^{\ell-1}))+\mathbf{Z}^{\ell-1}, &\ell=1 \ldots N_L, \\
        &\hat{\mathbf{Z}}^{\ell}=\bar{\alpha}^{\ell} \operatorname{MLP}(\operatorname{LN}(\overline{\mathbf{Z}}^{\ell}))+\overline{\mathbf{Z}}^{\ell},  &\ell=1 \ldots N_L,
    \end{aligned}
\end{equation}
where $N_L, \alpha^\ell$, and $\bar{\alpha}^{\ell}$ denote the number of enocder layers and learnable scaling factors per channel initialized to zeros \cite{goingdeepimageTrans}. We only add a 2D sine position embedding for a first encoder layer. The attention maps of the last encoder layer are fed into the head, which produces a temporally different contributed prediction in each branch, respectively. 

\subsection{Loss Function}

We design our loss function $\mathcal{L}$ as a sum of square error for a semi-supervised task and a prediction result: 
\begin{equation}
    \mathcal{L} = \mathcal{L}_{occ} + \mathcal{L}_{pred}.
\end{equation}
Here, the standard pose estimation loss is employed for each component. It can be formulated as:
\begin{equation}
    \begin{aligned}
    \mathcal{L}_{occ} & = \frac{1}{N_j} \sum_{j=1}^{N} v_{j} \|\hat{G}_j-\psi_{total,j}\|_{2}^2,\\
    \mathcal{L}_{pred} & =\frac{N_f}{N_j} \sum_{j=1}^{N} v_{j} \|G_j-P_j\|_{2}^2, 
    \end{aligned}
\end{equation}
where $P_j$ and $v_{j}$ denote prediction and visibility for joint $j$, respectively. The first loss term $\mathcal{L}_{occ}$ performs the supervision of occlusion-aware heatmaps, and the second loss term $\mathcal{L}_{pred}$ penalizes the difference between prediction and ground truth. We give more penalties to $\mathcal{L}_{pred}$ by giving a weight $N_f=8$ to further enhance the final prediction result. These Euclidean distance are valid when a joint $j$ is visible.

\begin{figure*}[t]
\centering
\includegraphics[width=0.9\linewidth]{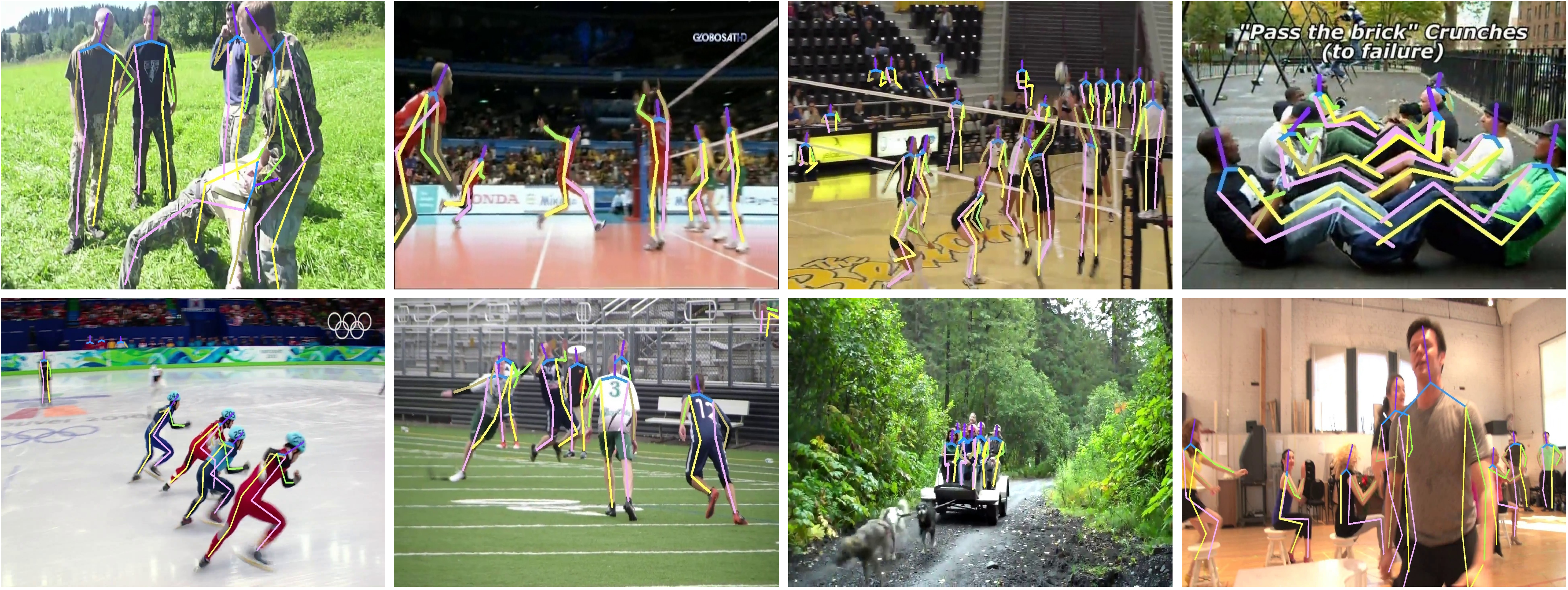}
\caption{Qualitative results on PoseTrack 2018 test set. Our model performs well in complex scenes such as occlusion, fast motion, and multiple humans.}
\label{fig:Qualitative results of complex scenes on PoseTrack 2018 test set.}
\end{figure*}

\setlength{\tabcolsep}{3pt}
\begin{table}[t]
\begin{center}
\caption{Quantitative Results on PoseTrack2017 validation set.}
\label{table:2017 validation set}
{\scriptsize
\begin{tabular}{l|ccccccc|c}
\hline
Method         & Head & Shoulder & Elbow & Wrist & Hip & Knee & Ankle & Mean\\ \hline
PoseFlow \cite{poseflow} & 66.7 & 73.3 & 68.3 & 61.1 & 67.5 & 67.0 & 61.3 & 66.5 \\
JointFlow \cite{jointflow} & -    & -    & -    & -    & -    & -    & -    & 69.3 \\
SimpleBaseline \cite{simplebaseline} & 81.7 & 83.4 & 80.0 & 72.4 & 75.3 & 74.8 & 67.1 & 76.7 \\
HRNet \cite{hrnet} & 82.1 & 83.6 & 80.4 & 73.3 & 75.5 & 75.3 & 68.5 & 77.3 \\
MDPN \cite{mdpn}  & 85.2 & 88.8 & 83.9 & 77.5 & 79.0 & 77.0 & 71.4 & 80.7 \\
PoseWarper \cite{posewarper} & 81.4 & 88.3 & 83.9 & 78.0 & 82.4 & 80.5 & 73.6 & 81.2 \\
DCPose \cite{dcpose} & 88.0 & 88.7 & 84.1 & 78.4 & 83.0 & 81.4 & 74.2 & 82.8 \\ \hline
\textbf{OTPose}  & \textbf{90.7} & \textbf{91.5} & \textbf{86.5} & \textbf{80.2} & \textbf{85.2} & \textbf{84.6} & \textbf{80.5} & \textbf{86.0} \\ \hline
\end{tabular}}
\end{center}
\end{table}

\setlength{\tabcolsep}{3pt}
\begin{table}[t]
\begin{center}
\caption{Quantitative Results on PoseTrack2017 test set.}
\label{table:2017 test set}
{\scriptsize
\begin{tabular}{l|ccccccc|c}
\hline
Method         & Head & Shoulder & Elbow & Wrist & Hip & Knee & Ankle & Mean\\ \hline
PoseFlow \cite{poseflow} & 64.9	& 67.5	& 65.0	& 59.0	& 62.5	& 62.8	& 57.9	& 63.0\\
JointFlow \cite{jointflow} & -	& -	& -	& 53.1	& -	& -	& 50.4	& 63.4\\
DetTrack \cite{dettrack} & -	& -	& -	& 69.8	& -	& -	& 65.9	& 74.1\\
SimpleBaseline \cite{simplebaseline} & 80.1	& 80.2	& 76.9	& 71.5	& 72.5	& 72.4	& 65.7	& 74.6\\
HRNet \cite{hrnet}	& 80.0	& 80.2	& 76.9	& 72.0	& 73.4	& 72.5	& 67.0	& 74.9\\
PoseWarper \cite{posewarper} & 79.5	& 84.3	& 80.1	& 75.8	& 77.6	& 76.8	& 70.8	& 77.9\\
DCPose \cite{dcpose} & 84.3	& 84.9	& 80.5	& 76.1	& 77.9	& 77.1	& 71.2	& 79.2\\\hline
\textbf{OTPose} & \textbf{85.1} & \textbf{85.3} & \textbf{81.0} & \textbf{76.2} & \textbf{78.5} & \textbf{77.7} & \textbf{72.0} & \textbf{79.7} \\ \hline

\end{tabular}}
\end{center}
\end{table}

\setlength{\tabcolsep}{3.7pt}
\begin{table}[t]
\begin{center}
\caption{Quantitative Results on PoseTrack2018 validation set.}
\label{table:2018 validation set}
{\scriptsize
\begin{tabular}{l|ccccccc|c}
\hline
Method    & Head & Shoulder & Elbow & Wrist & Hip  & Knee & Ankle & Mean \\ \hline
AlphaPose \cite{alphapose} & 63.9 & 78.7     & 77.4  & 71.0  & 73.7 & 73.0 & 69.7  & 71.9 \\
MDPN \cite{mdpn}    & 75.4 & 81.2     & 79.0  & 74.1  & 72.4 & 73.0 & 69.9  & 75.0 \\
PoseWarper \cite{posewarper} & 79.9 & 86.3   & 82.4  & 77.5  & 79.8 & 78.8 & 73.2  & 79.7 \\
DCPose \cite{dcpose}  & 84.0 & 86.6     & 82.7  & 78.0  & 80.4 & 79.3 & 73.8  & 80.9 \\ \hline
\textbf{OTPose} & \textbf{87.3} & \textbf{89.7} & \textbf{85.3} & \textbf{80.2} & \textbf{82.3} & \textbf{83.0} & \textbf{79.8} & \textbf{84.2}  \\ \hline
\end{tabular}}
\end{center}
\end{table}

\setlength{\tabcolsep}{2.4pt}
\begin{table}[t]
\begin{center}
\caption{Quantitative Results on PoseTrack2018 test set.}
\label{table:2018 test set}
{\scriptsize
\begin{tabular}{l|ccccccc|c}
\hline
Method    & Head & Shoulder & Elbow & Wrist & Hip  & Knee & Ankle & Mean \\ \hline
AlphaPose++ \cite{alphapose,alpha++} & -	& -	& -	& 66.2	& -	& -	& 65.0	& 67.6 \\
DetTrack \cite{dettrack}	& -	& -	& -	& 69.8	& -	& -	& 67.1	& 73.5 \\
MDPN \cite{mdpn}	& -	& -	& -	& 74.5	& -	& -	& 69.0	& 76.4 \\
PoseWarper \cite{posewarper} & 78.9	& \textbf{84.4}	& 80.9	& 76.8	& 75.6	& 77.5	& 71.8	& 78.0 \\
DCPose \cite{dcpose} & 82.8	& 84.0	& 80.8	& 77.2	& 76.1	& 77.6	& 72.3	& 79.0 \\ \hline
\textbf{OTPose} & \textbf{83.0} & 84.0 & \textbf{81.6} & \textbf{77.4} & \textbf{76.2} & \textbf{78.2} & \textbf{72.8} & \textbf{79.6} \\ \hline
\end{tabular}}
\end{center}
\end{table}

\section{Experiments}

\subsection{Datasets}
PoseTrack dataset is a large-scale benchmark for multi-person pose estimation and tracking in videos. PoseTrack2017 \cite{posetrack2017} contains 514 videos, including 66,374 frames, split into 250, 50, and 214 videos for training, validation, and test set, respectively. PoseTrack2018 \cite{posetrack2018} increases the number of videos to 1,138 videos with 153,615 pose annotations and divides them into 593, 170, and 375 videos for training, validation, and test set. Every 30 frames from the center are annotated for training videos. For validation and test videos, every fourth frame is annotated. The annotations include 15 body keypoint locations with visibility, a unique person ID, and a head bounding box for each instance. We evaluate our model only for visible joints with average precision (AP), an evaluation metric of the PoseTrack datasets. 

\subsection{Empirical Setting}

We conduct scaling, truncating, random rotation, and flipping for data augmentation during training. The bounding box for human $i$ is obtained from frame $t$ by a human detector, enlarged by 25\%, and used to crop the supplement frames. Our model processes five consecutive images because using more frames causes high memory complexity, and using fewer frames results in inaccurate prediction results. By default, we set $P_h=P_w=1$ for patch sizes to capture pixel-level relations. In multi-head self-attention, we set the number of heads as 1 for Occlusion Encoder and 2 for Temporal Encoder. We use the pretrained HRNet-W48 model as our backbone network. All parameters are initialized from Gaussian distribution $\sim \mathcal{N}(0,\,0.001^{2})$ with zero bias. With an optimizer AdamW, we set an initial learning rate as $1 \times 10^{-5}$ to warm up at 12 epochs, which decays to zero for the rest of 38 epochs in a cosine annealing manner. We use a batch size of 32 and train our model with 4 Nvidia A100 GPUs.

\subsection{Comparisons}

\begin{figure}[t]
\centering
\includegraphics[width=\linewidth]{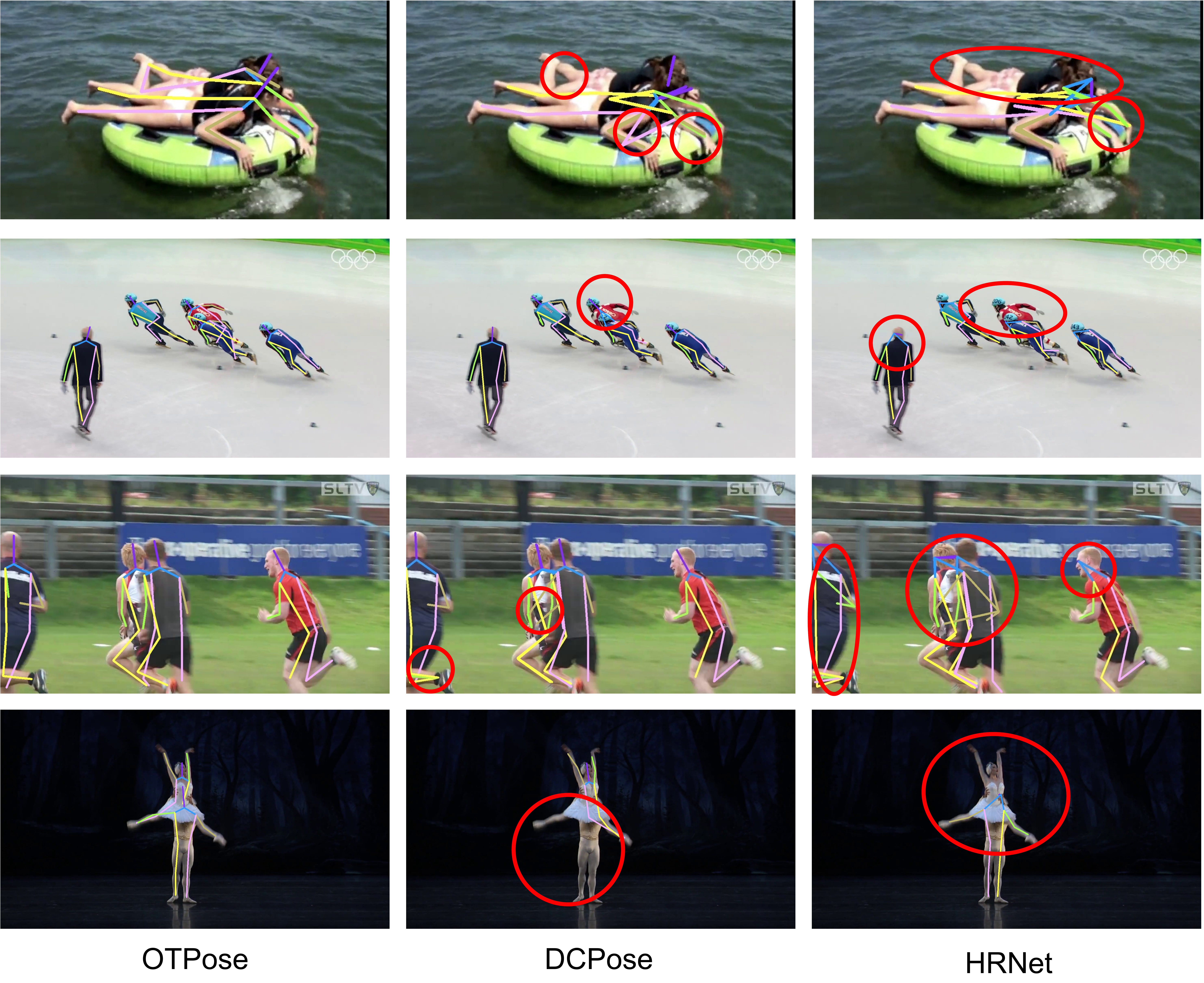}
\caption{Qualitative comparison on PoseTrack 2017 validation set. Since HRNet is an image-based method, it often fails, and DCPose produces relatively better prediction results. In contrast, our approach can estimate poses consistently. The solid red circles represent inaccurate predictions.}
\label{fig:Qualitative comparison of complex scenes on PoseTrack 2017 validation set.}
\end{figure}

\subsubsection{\textbf{Quantitative Results}}
To verify the effectiveness of our methods, we compare our approach against current state-of-the-art frameworks for pose estimation. Tables ~\ref{table:2017 validation set} and \ref{table:2017 test set} present the quantitative results on the validation set and test set of PoseTrack2017. We demonstrate that our method outperforms the existing approaches \cite{hrnet,mdpn,posewarper,dcpose,poseflow,jointflow,simplebaseline,dettrack}. We also represent the comparison results on the PoseTrack2018 dataset in Tables ~\ref{table:2018 validation set} and \ref{table:2018 test set}. As shown in the Tables, we once again indicate that our method performs better than current methods \cite{alphapose,mdpn,posewarper,dcpose,alpha++,dettrack}. 

\subsubsection{\textbf{Qualitative Results}}
Through Fig.~\ref{fig:Qualitative results of complex scenes on PoseTrack 2018 test set.}, we demonstrate that our method indicates remarkable results in complex scenes such as occlusion, crowded scenes, fast motion, and challenging poses. We follow previous methods for a precise comparison, using the same input size $384 \times 288$ and employing the same pretrained backbone network HRNet-W48 to process input images at test time. Our method is qualitatively validated by the comparisons with 1) DCPose \cite{dcpose} and 2) HRNet-W48 \cite{hrnet}. DCPose is a method that has long occupied state-of-the-art performance in the PoseTrack dataset by refining the pose of the current frame in the video using two-way frames. HRNet is a method that shows profound results for human pose estimation in images. This approach achieves superior performance by maintaining high-resolution representations through a multi-scale fusion of feature maps. We show that our method outperforms the existing methods for challenging cases. The comparison results are illustrated in Fig.~\ref{fig:Qualitative comparison of complex scenes on PoseTrack 2017 validation set.}.

\subsection{Ablation Study}
\subsubsection{\textbf{Effect of Each Component}}
We validate the effect of our component with an extensive ablation study. We conduct experiments on the PoseTrack2018 validation set and compare the results in Table~\ref{table:Ablation Study on PoseTrack2018 validation set}. Row 1 indicates a minimal version of our approach that removes one temporal branch (TB) and occlusion encoder (OE). The last row shows our complete model. In rows 2 and 3, we remove our components one by one from the full model to observe their contributions. The occlusion encoder (row 3) has a significant advantage in quantitative evaluation, followed by temporal branch (row 2). In row 3, the occlusion-aware heatmaps $\psi_{total}^i$ are replaced by occlusion attention masks $M_{total}$.
 
\subsubsection{\textbf{Effect of Occlusion Attention Mask}}
Additionally, rows 4 represents that the pseudo label's mask component is replaced. The pseudo label is created by replacing the occlusion attention mask $M_{total}$ with a pose difference $F_{total}$; observing that an explicit mask has benefits in improving the prediction results of overall keypoints, including occluded joints.

\subsubsection{\textbf{Scaling the Size of Encoders}}
We study how the performance depends on the size of the transformer encoder, as shown in rows 5-7. The number of layers increases from 5 to 8, and the performance increases consistently then decreases when we get more layers in the encoder. Further, we conduct attention between patch to validate our keypoint-wise attention mechanism in rows 8 and 9. To use high-resolution images $384\times288$, we adopt local self-attention \cite{localAttention}, window size 19 and 36, which has recently been validated for performance and efficiency. They indicate that keypoint-wise attention has an advantage quantitatively.

\setlength{\tabcolsep}{2pt}
\begin{table}[t]
\begin{center}
\caption{Ablation Study to observe the contribution of each component of our model on PoseTrack2018 validation set. $A \leftarrow B$ refers feature $B$ is replaced by $A$.}
{\scriptsize
\label{table:Ablation Study on PoseTrack2018 validation set}
\begin{tabular}{l|ccccccc|c}
\hline
Method & Head & Shoulder & Elbow & Wrist & Hip  & Knee & Ankle & Mean \\ \hline
1) one TB, w/o OE & 87.4 & 89.8 & 83.7 & 77.6 & 82.5 & 80.8 & 69.4 & 82.0 \\ 
2) one TB & 86.7 & 88.8 & 84.5 & 78.9 & 79.0 & 80.7 & 78.9 & 82.8  \\
3) w/o OE $\psi \longleftarrow  M$ & 87.9 & 88.1 & 82.4 & 76.5 & 81.3 & 80.9 & 75.3 & 82.2 \\ \hline
4) Pseudo label $ M \leftarrow  F$ & 85.4 & 89.0 & 84.6 & 78.3 & 81.4 & 82.2 & 78.3 & 82.9 \\ \hline
5) \# $N_L$ = 5 & 86.6 & 89.3 & 84.9 & 79.3 & 81.2 & 82.7 & 79.5 & 83.6 \\
6) \# $N_L$ = 6 & 87.2 & 89.0 & 84.7 & 79.6 & 79.2 & 81.4 & 79.7 & 83.3 \\ 
7) \# $N_L$ = 7 & 88.6 & \textbf{90.3} & 84.6 & 78.3 & \textbf{84.1} & 82.7 & 77.9 & 84.1 \\ \hline
8) Window size = 19 & 89.0 & 90.2  & 83.4 & 77.5 & 82.8 & 81.7 & 77.3 & 83.5 \\
9) Window size = 36 & 87.1 & 89.0    & 83.7 & 77.5  & 82.9 & 82.8 & 77.0 & 83.1 \\ \hline
10) \textbf{OTPose}, \# $N_L$ = 8 & \textbf{87.3} & 89.7 & \textbf{85.3} & \textbf{80.2} & 82.3 & \textbf{83.0} & \textbf{79.8} & \textbf{84.2}  \\ \hline
\end{tabular}}
\end{center}
\end{table}

\section{Conclusion}
In this work, we directly addressed occlusion issues and aggregated temporal features by constructing a transformer-based model to effectively exploit sparsely annotated videos. We demonstrated that explicitly learning occlusion in a semi-supervised manner enables detecting easily occluded keypoints, such as wrists and ankles, as well as overall joints. Our framework OTPose achieved state-of-the-art performance on the Posetrack2017 \cite{posetrack2017} and Posetrack2018 datasets \cite{posetrack2018} and reduced the need for dense annotations. We are interested in reviewing our task in a self-supervised manner, which can further reduce the need for annotations by combining it with our approach.

\bibliography{mybib.bib}
\bibliographystyle{IEEEtran}

\end{document}